\documentclass[journal]{IEEEtran}
\usepackage{booktabs} 
\usepackage{array}
\usepackage{multirow}
\usepackage{cite}
\usepackage{graphicx}
\usepackage{subcaption}
\usepackage[font=small]{caption}
\usepackage{hyperref}
\usepackage{bm}
\usepackage{amsfonts}
\usepackage{amsmath}
\usepackage{amssymb}
\usepackage{siunitx}
\usepackage{algorithm}
\usepackage{algorithmic}

\begin{document}
\title{End-to-End Human Pose Reconstruction from Wearable Sensors for 6G Extended Reality Systems}
\author{
    Nguyen Quang Hieu, Dinh Thai Hoang, Diep N. Nguyen, \\Mohammad Abu Alsheikh, Carlos C. N. Kuhn, Yibeltal F. Alem, and Ibrahim Radwan 
    \thanks{N. Q. Hieu is with School of Electrical Data Engineering, University of Technology, Sydney, NSW 2007, Australia, and also with School of Information Technology and Systems, University of Canberra, Canberra, ACT 2617, Australia, emails: (hieu.nguyen-1@student.uts.edu.au; hieu.nguyen@canberra.edu.au).}
    \thanks{D. T. Hoang, and D. N. Nguyen are with the School of Electrical and Data Engineering, University of Technology Sydney, NSW 2007, Australia, emails: (diep.nguyen@uts.edu.au; hoang.dinh@uts.edu.au).}
    \thanks{M. A. Alsheikh, C. C. N. Kuhn, Y. F. Alem, and I. Radwan are with Faculty of Science \& Technology, University of Canberra, Canberra, ACT 2617, Australia, emails: \{Mohammad.Abualsheikh; Carlos.NoschangKuhn; Yibe.Alem; Ibrahim.Radwan\}@canberra.edu.au.}
}
\maketitle
\begin{abstract}
Full 3D human pose reconstruction is a critical enabler for extended reality (XR) applications in future sixth generation (6G) networks, supporting immersive interactions in gaming, virtual meetings, and remote collaboration. However, achieving accurate pose reconstruction over wireless networks remains challenging due to channel impairments, bit errors, and quantization effects. Existing approaches often assume error-free transmission in indoor settings, limiting their applicability to real-world scenarios. 
To address these challenges, we propose a novel deep learning-based framework for human pose reconstruction over orthogonal frequency-division multiplexing (OFDM) systems. The framework introduces a two-stage deep learning receiver: the first stage jointly estimates the wireless channel and decodes OFDM symbols, and the second stage maps the received sensor signals to full 3D body poses.
Simulation results demonstrate that the proposed neural receiver reduces bit error rate (BER), thus gaining a 5 dB gap at $10^{-4}$ BER, compared to the baseline method that employs separate signal detection steps, i.e., least squares channel estimation and linear minimum mean square error equalization. Additionally, our empirical findings show that 8-bit quantization is sufficient for accurate pose reconstruction, achieving a mean squared error of  $5\times10^{-4}$ for reconstructed sensor signals, and reducing joint angular error by 37\% for the reconstructed human poses compared to the baseline. 
\end{abstract}

\section{Introduction}
\subsection{Motivation and Related Works}
The extended reality (XR) use cases in the future communication systems, i.e., 6G, have attracted attention from researchers in recent years, driven by advancements in communication technologies and the promise of applications such as immersive gaming, virtual meetings, and augmented industrial training \cite{wang2023road, sebire2023extended, siriwardhana2021survey, zhu2024human}. 
XR devices are expected to bring emerging applications to users over wireless connectivity and 6G infrastructures. The challenges remain in optimizing the benefits of these XR devices, including addressing technical limitations such as latency, bandwidth constraints, and the energy efficiency of both devices and network infrastructure \cite{wang2023road}. 
One potential approach is to utilize sensing information from such devices and creating a feedback loop of sensing information and quality of experience.
This approach makes it possible to gain insightful information from XR devices for designing innovative applications \cite{wang2023road, zhu2024human}. Specifically, information such as the orientation of the inertial measurement units inside the XR devices is crucial for optimizing the quality of service at the base station, as it enables precise beamforming \cite{guerra2018single}, reduces latency, and ensures reliable communication links \cite{bjornson2019massive}. This enables more efficient wireless resource management and contributes to a seamless and immersive user experience.

There is a rich literature on utilizing sensing information from a single wearable sensor, i.e., one inertial measurement unit (IMU) sensor embedded inside the XR headset, to improve quality of service, e.g., \cite{zhang2019drl360}, \cite{hieu2023virtual}, \cite{perfecto2020taming}, and references therein. This line of work has been highly optimized for single head-mounted devices like virtual reality and augmented reality headsets/glasses. However, the use of a single sensor in these works limit their potential in scenarios where only the information about the user's head orientation is needed. For example, in emerging use cases such as virtual gaming and robotics, having more controllable sensors and actuators is always preferable, yet requiring more complex signal processing approaches. In these use cases, users may provide sensing information of the headset/glasses, joy-stick controllers, and some additional body joints' sensors \cite{winkler2022questsim}. By combining readings from all sensors on the human body, one can fully unleash innovative applications with a high degree of freedom in 3D space. This has been long studied in computer vision and robotic applications \cite{liu2022recent}. However, the direction is still new in the future wireless communication use cases. For example, how to use this sensing information over 6G wireless systems, e.g., using multiple antenna arrays with OFDM configurations, to enable innovative use cases, e.g., full 3D human pose reconstruction, remain an unanswered question.

In XR applications, reconstructing full-body poses in real-time is essential for rendering realistic avatars or enabling collaborative virtual environments. However, existing methods in the literature, e.g., \cite{huang2018deep, guzov2021human}, are limited in error-free settings, meaning that there is no packet lost and the learning model operates under ideal conditions. In our recent work in \cite{hieu2024reconstructing}, we explored the human pose reconstruction in wireless systems in a generic and simple additive white Gaussian noise (AWGN) channel model. Our initial findings in \cite{hieu2024reconstructing} demonstrated the impacts of the AWGN channel on the final human pose reconstruction result at the receiver. However, the use of the AWGN channel is far from realistic in industrial settings with OFDM systems and multiple-input-multiple-output (MIMO) antenna arrays.

\subsection{Challenges and Proposed Solutions}
In this work, we aim to bridge the gap between human pose reconstruction using wearable sensors and OFDM channel estimation at the wireless base station. On the one hand, learning-based approaches can be utilized to reconstruct full 3D body poses and movements from IMU sensors, e.g., \cite{von2017sparse, huang2018deep, guzov2021human}. On the other hand, the receiver needs to account for the relationship between bit error rate (BER) values and the quality and robustness of the final pose reconstruction. This is because BER and packet loss directly distort IMU signals, resulting in degraded pose reconstruction at the receiver. Additionally, transmitting high-precision raw IMU signals, i.e., using a 32-bit floating-point data type, can impose a significant burden on the uplink channel. Thus, data quantization with a lower number of bits, e.g., 8 bits or 10 bits, is necessary in the source coding process. However, the effects of quantization on the final reconstructed signals and body poses remain unexplored in the literature. This highlights the need for channel estimation and signal decoding at the receiver, with the goal of minimizing BER.

The research gaps mentioned above motivate us to propose an end-to-end framework for human pose reconstruction over future wireless systems, targeting XR use cases in 6G. To address the challenges of bit errors, quantization effects, and channel impairments, our framework leverages a two-stage deep learning-based receiver approach. In the first stage, the receiver jointly estimates the wireless channel and decodes OFDM symbols, minimizing the BER and ensuring reliable signal transmission. In the second stage, the decoded IMU signals are mapped to full 3D human body poses using a deep neural network, even in the presence of quantization and packet errors. Our approach is novel because it integrates channel estimation, signal decoding, and pose reconstruction into a single framework, enabling robust and accurate human pose reconstruction under realistic wireless conditions.

The significance of our approach lies in its ability to handle low-bit quantization (e.g., 8-bit precision) without sacrificing reconstruction accuracy, reducing the uplink channel burden while maintaining high fidelity of reconstructed human poses. Furthermore, our framework is designed to operate in site-specific OFDM channels, modeled using ray tracing, which captures real-world propagation effects such as multipath and Doppler shifts. This makes our solution highly applicable to practical XR scenarios in 6G networks, where reliable communication is critical for immersive experiences. Overall, our main contributions in this work are as follows:
\begin{itemize}
\item \textbf{End-to-end framework for human pose reconstruction over OFDM systems}:
We propose a novel framework that integrates human body pose reconstruction with OFDM-based wireless communication. This is the first end-to-end system model designed to align with practical constraints in XR use cases. The key design principle of our approach is a two-stage deep learning receiver that consists of (i) an OFDM receiver for channel estimation and OFDM symbol decoding and (ii) an IMU receiver for mapping between decoded IMU signals into 3D human poses.

\item \textbf{Neural receiver for joint channel estimation and signal decoding}:
We introduce a neural receiver at the base station that jointly estimates the wireless channel and decodes OFDM symbols. Unlike conventional OFDM receivers that rely on multi-step signal processing, our approach leverages joint training of channel estimation, equalization, and demodulation to learn an optimal signal decoding strategy. We empirically show that this joint training is equivalent to minimizing the residual information in multi-step signal processing approaches.

\item \textbf{Mapping received IMU signals to 3D human body poses}:
We propose a simple yet effective method for mapping received IMU signals at the receiver to specific human body poses. This is achieved by training a deep neural network with ground-truth human poses as labels and IMU signals as inputs. We demonstrate that, even in the presence of data quantization and packet errors during transmission, human poses can be reconstructed with high accuracy. Specifically, we achieve a mean squared error (MSE) of $1.8 \times 10^{-4}$ for received sensor signals under a signal-to-noise power ratio of $E_b/N_0 = 5$ dB using a low quantization level of 6 bits.

\item \textbf{Experimental validation of the proposed framework}:
Our experiments validate the superiority of the proposed framework over conventional receivers that employ least squares channel estimation and linear minimum mean squared error equalization. The results demonstrate smoother pose animations under realistic $E_b/N_0$ conditions. These findings highlight the critical interplay between wireless channel reliability and XR immersion, advancing the vision of seamless human-machine interaction in future wireless communication systems. To further support open research and reproducibility, we release the code to generate all results in this paper at \url{https://github.com/TheOpenSI/imu2pose-sionna}.
\end{itemize}

The rest of the paper is organized as follows. Section \ref{sec:system-model} describes the system model of the proposed framework. Section \ref{sec:simulation} reports simulation results and findings. Finally, Section \ref{sec:conclusion} concludes the paper. 
Throughout the paper, we use the following mathematical notations. Scalars are represented by lowercase italic letters, e.g., $x \in \mathbb{R}$. Vectors are denoted by bold lowercase letters, e.g., $\bm{x} \in \mathbb{R}^N$, while matrices are represented by bold uppercase letters, e.g., $\mathbf{X} \in \mathbb{R}^{M \times N}$. The Euclidean (L2) norm of a vector $\bm{x}$ is defined as $\|\bm{x}\|_2 = \sqrt{\sum_{i=1}^{N} x_i^2}$, whereas the L1 norm is given by $\|\bm{x}\|_1 = \sum_{i=1}^{N} |x_i|$. 

\section{System Model}
\label{sec:system-model}
\begin{figure*}[t]
\centering
\includegraphics[width=0.9\linewidth]{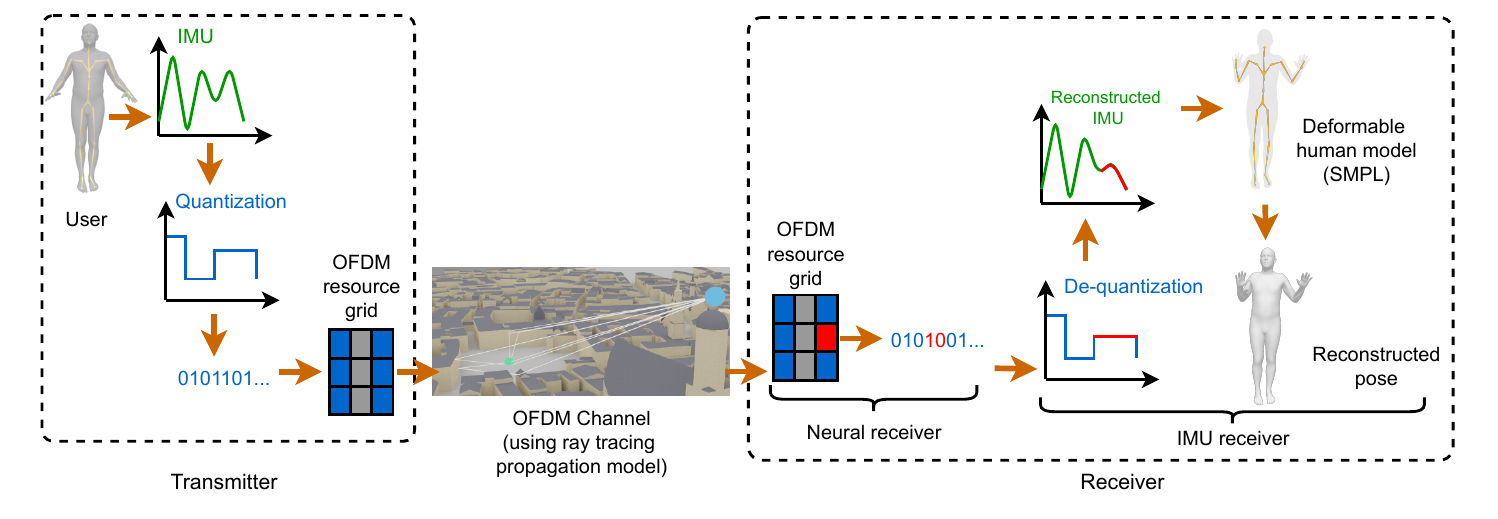}
\caption{Our single-user OFDM system in which the user is equipped with a single antenna placed at the XR headset \cite{duru2024pose}. The transmitter sends IMU signals over the uplink channel to the receiver. The IMU signals are quantized before transmission over the OFDM channel. The OFDM channel is modelled with a ray tracing propagation method in Sionna library \cite{sionna}. Finally, the receiver decodes the OFDM symbols into IMU signals and then maps the IMU signals into specific human body poses.}
\label{fig:system-model}
\end{figure*}

Fig.~\ref{fig:system-model} illustrates an end-to-end system model that can be divided into three main building blocks that are transmitter, OFDM channel, and receiver. As illustrated in Fig.~\ref{fig:system-model}, IMU signals are quantized before being transmitted over an OFDM channel. 
After quantization and modulation, the symbols are transmitted over a site-specific OFDM channel modelled via ray tracing. At the receiver, the signals are detected and decoded. Due to potential bit errors (denoted as red binary digits in Fig.~\ref{fig:system-model}), the reconstructed IMU signals can be distorted. The distortion of the IMU signals over the OFDM channel results in the degraded quality of the reconstructed human poses.

In the following, we describe the main building blocks in greater detail. After that, we describe the novel design in the system model, including the deep learning based OFDM receiver, denoted as ``Neural receiver" in Fig.~\ref{fig:system-model}, that can learn the channel estimation and signal decoding at the same time. In addition, we propose a novel method for transforming the received IMU signals into specific human poses in 3D, denoted as ``IMU receiver" in Fig.~\ref{fig:system-model}, thus enabling various downstream applications in XR scenarios.

\subsection{Transmitter}
The transmitter is a head-mounted device, e.g., an XR headset, of the user and is equipped with an antenna array \cite{duru2024pose}. In this work, the transmitter can transmit the IMU signals measuring the orientation and acceleration of the user's body joints, e.g., hands, elbows, legs,  hip, and head. For this, the user's body is attached with wearable sensors (i.e., IMUs). The signals measured at the sensors are assumed to be synchronized and ready to be transmitted at the head-mounted device. In our work, the user is equipped with 17 IMU sensors, similar to the works in literature \cite{von2017sparse}, \cite{huang2018deep} and \cite{hieu2024reconstructing}. The signal processing pipeline at the transmitter is as follows.

\subsubsection{Signal quantization}
The main goal of signal quantization at our transmitter is twofold. First, it converts the high dimensional continuous data like IMU signal into the binary sequence that is ready for signal modulation and transmission. Second, signal quantization also removes redundant signals by using potential lower number of bits. For example, the original measurement of the IMU sensor is usually in 32 floating-bit precision, meaning that one needs 32 bits (4 bytes) to represent a single data point, e.g., one orientation feature, within the IMU sensor reading. However, the use of raw precision requires a significant amount of data redundancy. For example, as many of the IMU sequences contain low-frequency coefficient values \cite{hieu2024reconstructing}, e.g., when the user stands still or moves slowly, the use of 32-bit precision is highly redundant. As a result, the quantization process at the transmitter is critical for transforming high-precision data samples into binary sequences with lower redundancy. 

In our work, we employ a simple uniform quantization process by dividing each IMU data feature into $2^q$ intervals, where $q$ is the precision of the quantization. In other words, each IMU feature in the original interval $[-1.0, 1.0]$ is mapped into a binary sequence with length $q$ bits, where the lowest value $-1.0$ can be represented as $q$ zero-valued bits and the highest value $1.0$ can be represented as $q$ one-valued bits. Let $\bm{x} \in \mathbb{R}^{204}$ denote the original IMU data frame of the 204-dimension vector, which is collected from 17 wearable IMU sensors  \cite{huang2018deep, hieu2024reconstructing}. Each IMU feature $x_k \in [-1.0, 1.0]$ within  $\bm{x}$ is quantized into $q$ binary bits as follows:
\begin{equation}
x_k \mapsto [b_{(k-1)q + 1}, b_{(k-1)q + 2}, \dots, b_{kq}], \quad k = 1, 2, \dots, 204.
\label{eq:quantization}
\end{equation}

From hereafter, we denote the binary sequence representing of $\bm{x} \in \mathbb{R}^{204}$ with $q$-bit quantization as: 
\begin{equation}
\bm{b} = [b_1, b_2, \dots, b_{204 \cdot q}], \quad b_k \in \{0, 1\}, \; k = 1, 2, \dots, 204 \times q.
\end{equation}

The above uniform quantization is simple and easy to implement, yet provides useful information about the redundancy in the data. In particular, we find that using $q=8$ bits is enough to fully capture the information of the IMU data with minimal data distortion. Any values that are larger than $8$-bit may cause redundancy. More details and analyses will be discussed in Section \ref{sec:simulation}.

\subsubsection{Signal modulation and transmission}
\begin{figure*}[t]
\centering
\includegraphics[width=0.8\linewidth]{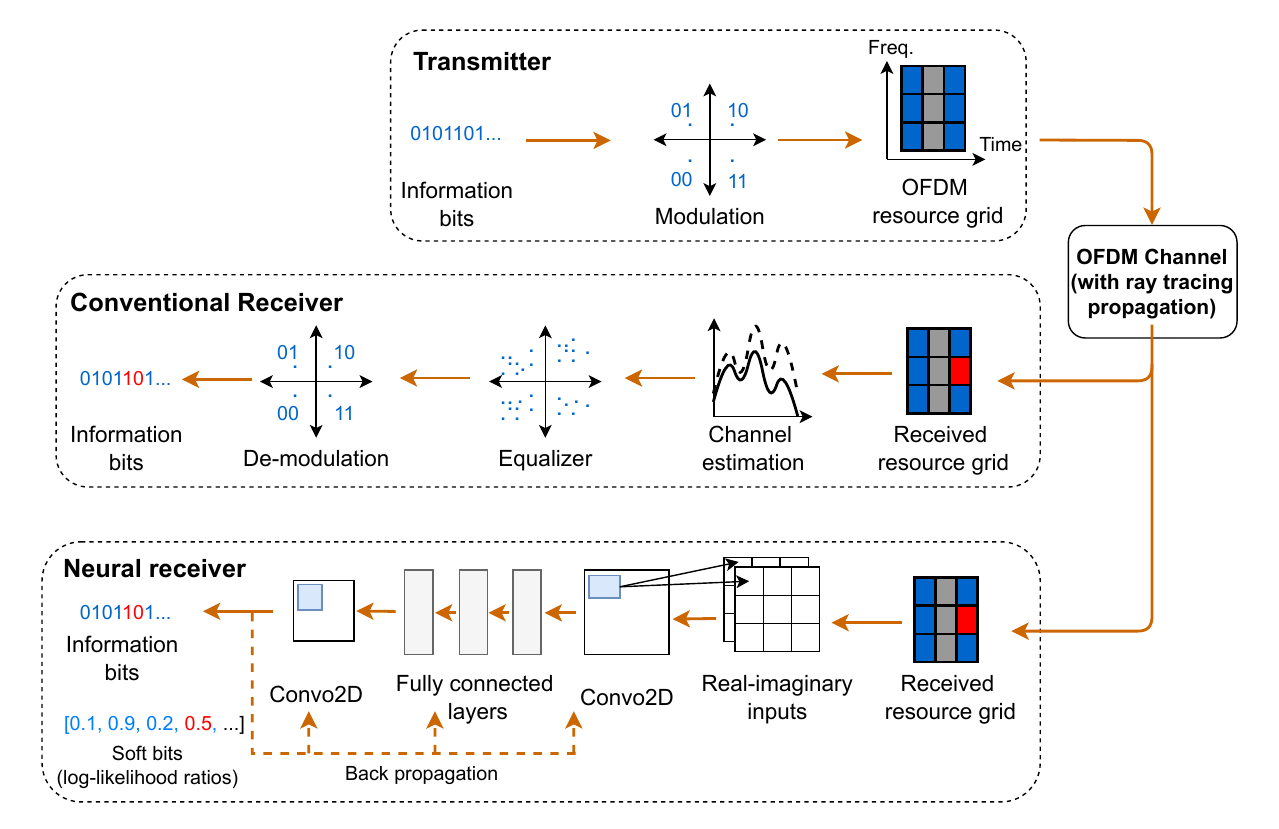}
\caption{Neural receiver approach to decode the received OFDM resource grid into information bits. Unlike conventional receiver that requires multiple processing blocks, i.e., channel estimation, equalization, and demodulation, the neural receiver approach jointly learns the parameters through supervised learning. At test time, the pre-trained neural receiver is capable of performing signal detection in real-time.}
\label{fig:neural-receiver}
\end{figure*}

After the original data $\bm{x}$ is transformed into the binary sequence $\bm{b}$, we need to perform modulation, i.e., transforming binary sequence $\bm{b}$ into complex numbers, i.e., symbols, denoted as $\bm{c}$. 
Each symbol is a single point in an Euclidean space, representing a group of a fixed number of bits. In this work, we follow the standard quadrature amplitude modulation (QAM) scheme, as illustrated in the ``Transmitter'' block of Fig.~\ref{fig:neural-receiver}. After modulation, the symbols are mapped onto the OFDM resource grid, in which each available bandwidth is divided into multiple orthogonal subcarriers, each carrying a portion of the modulated symbols. In addition to the data resource blocks (illustrated as blue color blocks in Fig.~\ref{fig:neural-receiver}), there are pilot symbols inserted for channel estimation at the receiver (illustrated as gray color blocks in Fig.~\ref{fig:neural-receiver}). Once the OFDM resource grid is ready, we can transmit the data and pilot blocks via the OFDM channel. The OFDM channel modeling is described in the following.

\subsection{Channel Modelling with Ray Tracing}
\label{subsec:rt-introduction}
\begin{figure}[t]
    \centering
    \begin{subfigure}[t]{0.51\linewidth}
        \centering
        \includegraphics[width=\linewidth]{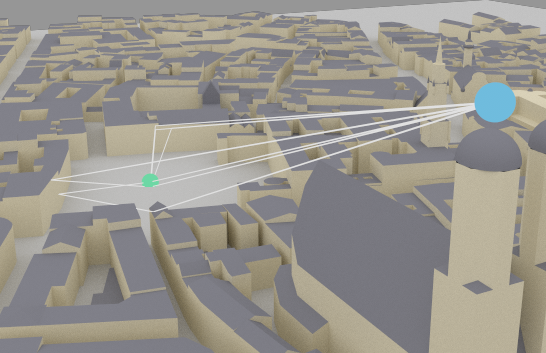}
        \caption{}
    \end{subfigure}
    \hfill
    \begin{subfigure}[t]{0.47\linewidth}
        \centering
        \includegraphics[width=\linewidth]{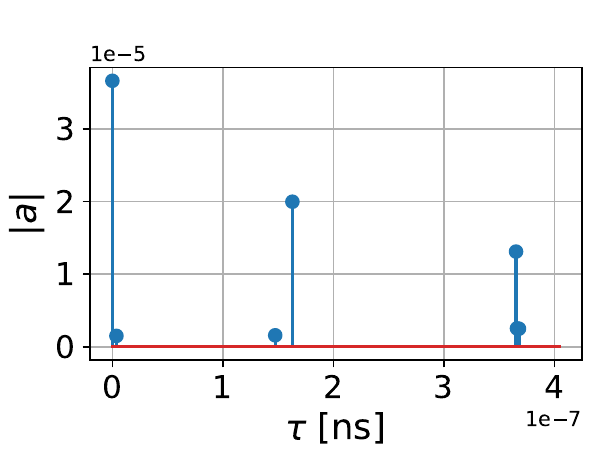}
        \caption{}
    \end{subfigure}
    \caption{(a) A 3D map of an area in Munich (Germany) with propagation paths created using the Sionna Ray Tracing toolkit. Figure (b) is the channel impulse response realization of the paths in figure (a).}
    \label{fig:rt-scenes}
\end{figure}
Unlike conventional approaches that use statistically random multipath propagation environments, such as fading channels like Rayleigh and Nakagami-m, we utilize a ray tracing model to simulate the propagation of the  OFDM signals in actual 3D scenes \cite{hoydis2024learning}. In conventional statistical models, the amplitudes of the multipath components can be modelled as independent and identically distributed random variables. However, these approaches do not account for site-specific details, such as the physical layout of buildings, materials, or specific propagation paths. Motivated by the essential of capturing the deterministic effects of geometry and materials in actual scenarios such as urban areas or indoor environments, ray tracing is a promising approach for modelling advanced communication systems like millimeter-wave (mmWave) and terahertz bands, where line-of-sight (LoS) and deterministic reflections are dominant.

In Fig.~\ref{fig:rt-scenes}, we use two actual 3D scenes from the Sionna ray tracing toolkit to create a channel dataset of channel impulse responses (CIRs) in the considered OFDM system. For example, the scenes in Fig.~\ref{fig:rt-scenes}(a) and Fig.~\ref{fig:rt-scenes}(b) illustrate 14 propagation paths from the transmitter (light blue dot) to the receiver (green dot). We observe that depending on the geometric properties of the scene, i.e., building positions and reflecting materials, the propagation paths can be different, resulting in different channel impulse responses. Let $L$ denote the number of propagation paths between the transmitter and the receiver at carrier frequency $f$, and the channel frequency response $H(f)$ of the channel is computed by \cite{hoydis2024learning}:
\begin{equation}
\label{eq:cfr}
H(f) = \sum_{i=1}^{L} \underbrace{\frac{\lambda}{4\pi} \mathbf{C}_R\left(\theta_i^R, \phi_i^R\right)^\mathbf{H} \mathbf{T}_i \mathbf{C}_T\left(\theta_i^T, \phi_i^T\right)}_{\triangleq a_i} e^{-j 2 \pi f \tau_i},
\end{equation}
where $\lambda$ is the wavelength, $(\theta_i^T, \phi_i^T)$ and $(\theta_i^R, \phi_i^R)$ are the angles of departure (AoD) and angles of arrival (AoA) of the $i$-th path with delay $\tau_i$ and transfer function $\mathbf{T}_i: \mathbb{C}^2 \mapsto \mathbb{C}^2$. $\mathbf{C}_T(\theta_i^T, \phi_i^T) \in \mathbb{C}^2$ and $\mathbf{C}_R(\theta_i^R, \phi_i^R) \in \mathbb{C}^2$ are the transmit-antenna and receive-antenna patterns, respectively \cite{hoydis2024learning}.
For an OFDM system with $N$ even subcarriers spread $\delta_f$ apart, the channel is characterized by the discrete channel frequency response $H[n]$:
\begin{equation}
\label{eq:cfr-ofdm}
H[n] = H(f + n \delta_f), n = -\frac{N}{2}, \dots, \frac{N}{2}-1.
\end{equation}

In our later simulation, we create the entire OFDM channel model in realistic scenarios by using ray tracing to create the propagation paths, characterized by the AoD and AoA information with specific complex antenna patterns $\mathbf{C}_R(\theta_i^R, \phi_i^R)$ and $\mathbf{C}_T(\theta_i^T, \phi_i^T)$. As illustrated in Fig.~\ref{fig:rt-scenes}, the AoD and AoA information of the propagation paths depend on the positions of the transmitter-receiver and the layout and reflecting properties of the buildings in the scene.

\subsection{Conventional Receiver}
\begin{figure}
\centering
\includegraphics[width=1.0\linewidth]{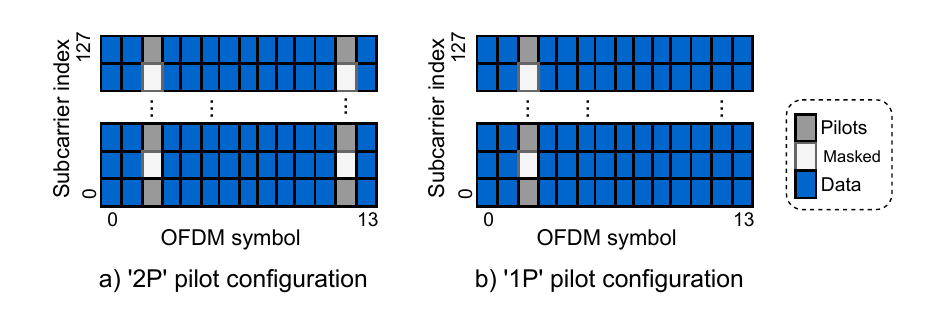}
\caption{Illustration of two pilot configurations,``2P" and ``1P", in an OFDM resource grid. The 2P configuration uses two pilot OFDM symbols (indices 2 and 12) for channel estimation, while the 1P configuration uses a single pilot OFDM symbol (index 2) \cite{aoudia2021end}. The resource grid also includes data subcarriers and masked subcarriers, as indicated in the legend.}
\label{fig:pilot-config}
\end{figure}

Before discussing our approach for the neural receiver and the IMU receiver, we first describe the conventional receiver in the OFDM system. The conventional signal processing pipeline is illustrated in Fig.~\ref{fig:neural-receiver} in the ``Receiver" block. The input of the receiver is the received OFDM resource grid. The received signal at the received OFDM resource grid can be expressed as
\begin{equation}
\mathbf{y}_{ij} = \mathbf{H}_{ij} c_{ij} + \mathbf{e}_{ij},
\end{equation}
where $\mathbf{y}_{ij} \in \mathbb{C}^{N_r \times 1}$, $c_{ij} \in \mathbb{C}$, and $\mathbf{e}_{ij} \in \mathbb{C}^{N_r \times 1}$ represent the received signal, transmitted signal, and additive noise vectors, respectively. The indices $i$ and $j$ denote the OFDM symbol and subcarrier indices, respectively.  Here, $\mathbf{H}_{ij} \in \mathbb{C}^{N_r \times 1}$ represents the channel vector for $N_r$ receiver antennas at the $i$-th OFDM symbol and $j$-th subcarrier. Note that $H[n]$ in (\ref{eq:cfr-ofdm}) is a scalar value that represents the frequency-domain channel response for the $n$-th subcarrier and does not include spatial information. Therefore, $H[n]$ should not be confused with $\mathbf{H}_{ij}^{(k)}$, an element of $\mathbf{H}_{ij}$, which contains both the frequency-domain channel response for subcarrier n and the spatial channel response for the $k$-th receiver antenna.

By using a fixed pattern for pilot symbols, as illustrated in OFDM resource grid configuration in Fig.~\ref{fig:pilot-config}, the receiver can estimate the channel state information based on the received pilots. After that, the channel state information of the entire resource grid can be interpolated from the pilot symbol positions. These channel estimation and equalization can be described as follows.

The estimated channel $\mathbf{\hat{H}}_{ij}$ can be derived by using the least square (LS) channel method as follows \cite[Equation 2.254]{heath2018foundations}: 
\begin{equation}
\mathbf{\hat{H}}_{ij}= \mathbf{y}_{ij} c_{ij}^*, \forall (i, j) \in \mathcal{P},
\label{eq:ls-estimation}
\end{equation}
where $\mathcal{P}$ denotes the set of indices for the pilot symbols in the OFDM resource grid and $c_{ij}^*$ is the complex conjugate of $c_{ij}$, assuming normalized power for pilots $|c_{ij}|^2 = 1$. The multiplying by the complex conjugate $c_{ij}^*$ helps reverse the phase shift introduced by the channel. 

Given a specific pilot configuration, e.g., 2P configuration in Fig.~\ref{fig:pilot-config}(a), the channel state information at the pilot symbols are estimated through (\ref{eq:ls-estimation}). After that, the channel state information of the entire OFDM symbol needs to be interpolated from these pilot positions to the entire time-frequency resource grid. A linear minimum mean
square error (LMMSE) equalization technique can be used to interpolate from pilot symbols, resulting in estimated symbols of the entire resource grid \cite[Example 2.21]{heath2018foundations}:
\begin{equation}
\hat{c}_{ij} = \Big(\mathbf{\hat{H}}_{ij}^{H} \mathbf{\hat{H}}_{ij} + \hat{\sigma}_e^2 \mathbf{I}\Big)^{-1} \mathbf{\hat{H}}_{ij}^{H} \mathbf{y}_{ij},
(i, j) \in \mathcal{D},
\label{eq:lmmse-equalization}
\end{equation}
where $\mathcal{D}$ denotes the set of indices of the data symbols and subcarriers, $\hat{\sigma}_e^2$ is the estimated noise power, $\mathbf{I}$ is the identity matrix, and $(\cdot)^{H}$ is the Hermitian transpose.

Once the transmitted symbols $c_{ij}$ are estimated, the next step is to map these symbols to log-likelihood ratios (LLRs), which can be further hard-decoded or forwarded into an outer forward error correction code. For a given subcarrier $(i, j)$, the LLR value for the $l$-th bit of the transmitted symbol is computed as follows \cite{honkala2021deeprx}:
\begin{equation}
L_{ijl} = \log \Big(\frac{Pr(b_l=0 | \hat{c}_{ij})}{Pr(b_l=1 | \hat{c}_{ij})}\Big),
\label{eq:llr-lth-bit}
\end{equation}
where $Pr(b_l= 0 | \hat{c}_{ij})$ is the conditional probability that the transmitted bit is 0 given the observed symbol $\hat{c}_{ij}$, while $Pr(b_l= 1 | \hat{c}_{ij})$ refers to the transmitted bit is 1, and $l = 0, \dots, B - 1$ where $B$ is the number of bits per symbol. As we are considering the quadrature amplitude modulation (QAM) constellations, the LLR can be computed by comparing the Euclidean distance between the estimated symbol $\hat{c_{ij}}$ and the possible constellation points corresponding to each bit hypothesis. Given the noise is Gaussian with variance $\hat{\sigma}_e^2$, the LLR can be approximated as follows \cite{honkala2021deeprx}:
\begin{equation}
L_{ijl} \approx \frac{1}{\hat{\sigma}_e^2}\Big(\min_{c \in C_l^1}\|\hat{c}_{ij} - c\|_2^2 - \min_{c \in C_l^0}\|\hat{c}_{ij} - c\|_2^2\Big),
\end{equation}
where $C_l^0$ represents the constellation point for which $l$-th bit is 0, while $C_l^1$ represents the case $l$-th bit is 1. From hereafter, we refer to the conventional receiver using LS channel estimation and LMMSE equalization as ``LS-LMMSE" receiver to illustrate the multi-stage signal processing nature of this receiver.

\section{Proposed Two-stage Deep Learning-based Receiver}
\subsection{Neural Receiver}
\begin{figure*}[t]
\centering
\includegraphics[width=0.7\linewidth]{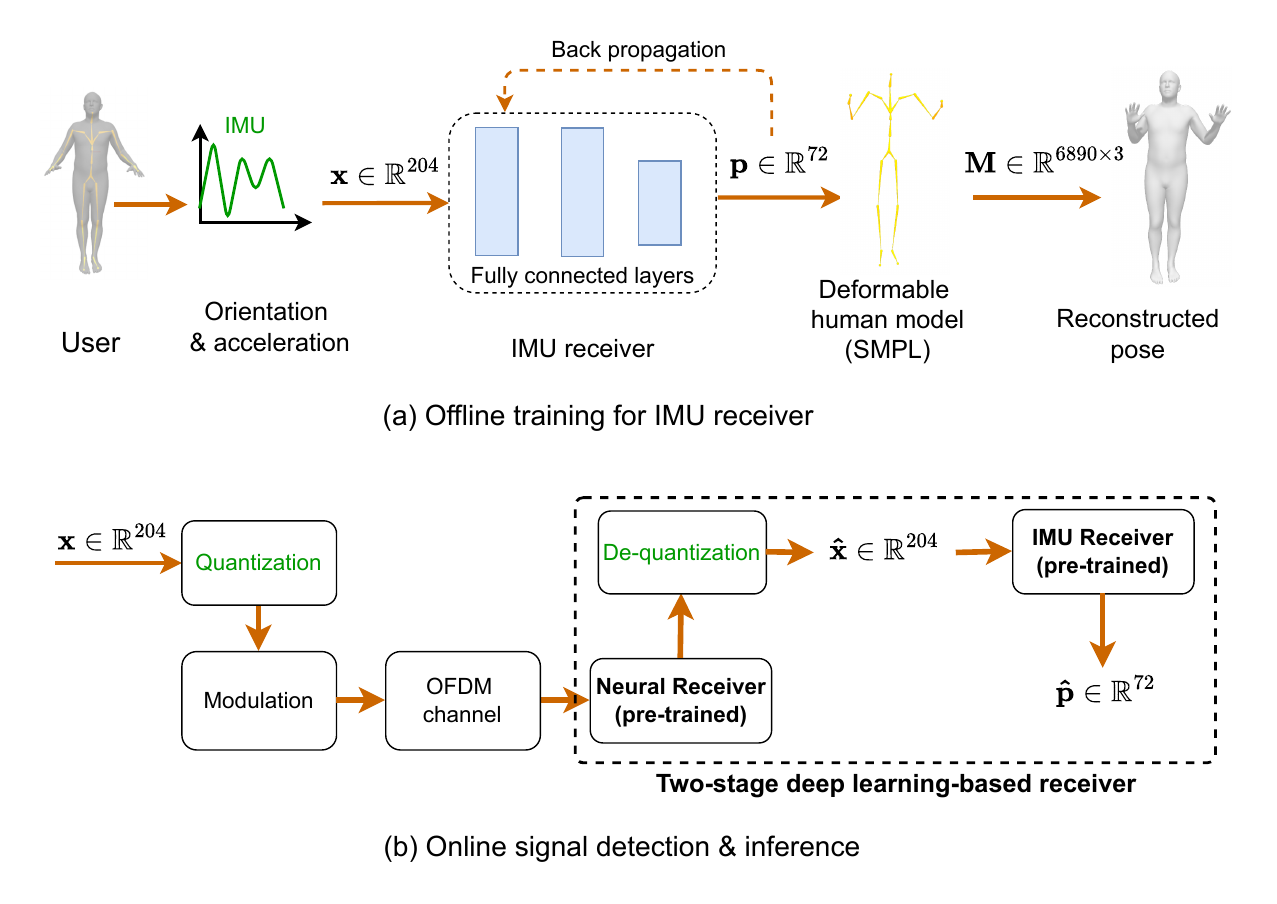}
\caption{Overview of the two-stage deep learning-based receiver for human pose reconstruction. (a) At the offline training phase, the IMU receiver is trained with raw IMU signals and ground truth human pose labels. (b) At the online signal detection and inference phase, the IMU signals are quantized and modulated before transmission. After that, the two-stage deep learning-based receiver perform simultaneous signal detection, de-quantization, and human pose prediction in real-time as no backpropagation occurs in this stage.}
    \label{fig:two_stage_receiver}
\label{fig:imu-receiver}
\end{figure*}

\begin{algorithm}[t]
\caption{Training Neural Receiver}
\label{algo:neural-receiver}
\begin{algorithmic}[1]
\REQUIRE Training dataset $\mathcal{D}_{\text{train}} = \{(\mathbf{y}_i, \mathbf{b}_i)\}_{i=1}^{N_{\text{train}}}$, noise variance $\sigma_e^2$, batch size $B_{\text{batch}} = 100$, number of iterations $N_{\text{iterations}} = 100,\!000$, learning rate $\eta = 0.001$.
\ENSURE Trained neural receiver $f_{\Theta}$.

\STATE Initialize neural receiver $f_{\Theta}$ with random weights $\bm{\Theta}$.
\FOR{iteration $= 1$ to $N_{\text{iterations}}$}
 		\STATE Sample noise variance $\sigma_e^2$ uniformly.
        \STATE Load batch: $\mathbf{y}_{\text{batch}} \in \mathbb{R}^{B_{\text{batch}} \times 100 \times 16 \times 14 \times 128}$, $\mathbf{b}_{\text{batch}} \in \{0, 1\}^{B_{\text{batch}} \times 204 \times q}$.
        \STATE Forward pass: Compute LLRs as in (\ref{eq:llr-lth-bit}) with estimated soft-bits as in (\ref{eq:estimated-soft-bits}).
        \STATE Compute loss as in (\ref{eq:loss-deeprx}).
        \STATE Backward pass: Compute gradients $\nabla_{\bm{\Theta}} \mathcal{L}_{\bm{\Theta}}$.
        \STATE Update weights: $\bm{\Theta} \leftarrow \bm{\Theta} - \eta \nabla_{\bm{\Theta}} \mathcal{L}_{\bm{\Theta}}$.
\ENDFOR
\RETURN Trained neural receiver $f_{\bm{\Theta}}$.
\end{algorithmic}
\end{algorithm}

As we observed from the multi-stage signal processing method in the conventional receiver, errors from early stages, e.g., LS channel estimation, can be propagated to the equalization steps, causing information loss. As each stage in the multi-stage signal processing can cause information loss, this results in reduced mutual information between the transmitted signals and the decoded signals. This means that the optimal mutual information between the transmitted symbol and the decoded symbol may not be fully preserved through all stages. The aforementioned limitations motivate an end-to-end approach with deep neural networks trained with a unified loss function, thus minimizing the information loss between conventional stages in LS channel estimation and LMMSE equalization \cite{honkala2021deeprx}.   

In this work, motivated by the success of end-to-end training for the receiver \cite{honkala2021deeprx, aoudia2021end},  we develop a deep learning model at the OFDM receiver, depicted as the neural receiver, to replace the traditional receiver. Our model design is illustrated in the ``Neural receiver" block in Fig.~\ref{fig:neural-receiver}. The input of the 2D convolutional (Conv2D) filter is the real and imaginary parts of the received OFDM resource grid. Following the Convo2D filter are the fully connected layers and the last output layer is another Convo2D filter. The output of the neural receiver is the log-likelihood ratio. We also use residual connections between the hidden layers and Convo2D layers to improve the robustness of the neural network to the large input datasets and deep layers \cite{he2016deep}. 

To train the neural receiver, we use the binary cross entropy as follows \cite{honkala2021deeprx}:
\begin{equation}
\mathcal{L}_{\bm{\Theta}} = -\frac{1}{|\mathcal{D}| B} \sum_{(i,j) \in \mathcal{D}} \sum_{l=0}^{B-1}\Big(b_{ijl} \log \hat{b}_{ijl} + (1 - b_{ijl}) \log(1 - \hat{b}_{ijl})) \Big),
\label{eq:loss-deeprx}
\end{equation}
where $|\mathcal{D}|$ is the total number of data symbols in the OFDM resource grid and $\hat{b}_{ijl}$ is the estimated probability that the bit $b_{ijl}$ is one, i.e.,
\begin{equation}
\hat{b}_{ijl} = \sigma(L_{ijl}) = \frac{1}{1 + e^{-L_{ijl}}},
\label{eq:estimated-soft-bits}
\end{equation}
in which $\sigma(\cdot)$ is the sigmoid function. 
The goal of training the neural receiver on the binary sigmoid cross-entropy loss is to optimize the receiver's ability to accurately estimate the transmitted bits from the noisy and distorted received signals, specifically by learning to output LLRs that maximize the likelihood of the correct bit decisions.
By optimizing the LLR outputs, the neural receiver minimizes the BER, leading to more reliable communication. Detailed parameters of the neural receiver are listed in Table \ref{tab:simulation-params}. 

The pseudo code for the training procedure of Algorithm \ref{algo:neural-receiver} can be described as follows. The algorithm begins by initializing the neural receiver model \(\bm{f_{\Theta}}\) with random weights \(\bm{\Theta}\). For each iteration, a batch of received OFDM resource grids \( \mathbf{y}_{\text{batch}} \) and corresponding ground-truth bits \( \mathbf{b}_{\text{batch}} \) are sampled. The forward pass computes log-likelihood ratios using (\ref{eq:llr-lth-bit}) and estimates soft bits using (\ref{eq:estimated-soft-bits}). The binary cross-entropy loss between the estimated and ground-truth bits is computed, and gradients are backpropagated to update the model weights using the Adam optimizer with learning rate \(\eta\). This process is repeated for \( N_{\text{iterations}} \) iterations, resulting in a trained neural receiver capable of robustly decoding OFDM symbols under realistic channel conditions. 

\subsection{IMU Receiver}
\begin{algorithm}[t]
\caption{Training IMU Receiver}.
\label{algo:imu-receiver}
\begin{algorithmic}[1]
\REQUIRE Training dataset $\mathcal{D}_{\text{train}} = \{(\bm{x}_i, \bm{p}_i)\}_{i=1}^{N_{\text{train}}}$, batch size $B_{\text{batch}}$, number of epochs $N_{\text{epochs}}$, learning rate $\eta$.
\STATE Initialize IMU receiver $f_{\bm{\Phi}}$ with random weights $\bm{\Phi}$.
\FOR{epoch $= 1$ to $N_{\text{epochs}}$}
    \STATE Shuffle training dataset $\mathcal{D}_{\text{train}}$ to ensure randomness.
    \FOR{batch $= 1$ to $\lceil N_{\text{train}} / B_{\text{batch}} \rceil$}
        \STATE Load batch: $\bm{x}_{\text{batch}} \in \mathbb{R}^{B_{\text{batch}} \times 204}$, $\bm{p}_{\text{batch}} \in \mathbb{R}^{B_{\text{batch}} \times 72}$.
        \STATE Forward pass:\\
         Compute predicted poses $\hat{\bm{p}}_{\text{batch}} = f_{\Phi}(\bm{x}_{\text{batch}})$.
        \STATE Compute loss as in (\ref{eq:loss-mse})
        \STATE Backward pass: \\
        Compute gradients $\bm{\Phi} \leftarrow \bm{\Phi} - \eta \nabla_{\bm{\Phi}} \mathcal{L}_{\text{MSE}}$.
        \STATE Update weights: $\bm{\Phi} \leftarrow \bm{\Phi} - \eta \nabla_{\bm{\Phi}} \mathcal{L}_{\text{MSE}}$.
    \ENDFOR
\ENDFOR
\RETURN Trained IMU receiver $f_{\Phi}$.
\end{algorithmic}
\end{algorithm}

The proposed IMU receiver and its role in the end-to-end system are illustrated in Fig.~\ref{fig:imu-receiver}.
In the offline training phase, i.e., Fig.~\ref{fig:imu-receiver}(a), the IMU signals produced from the set of IMU sensors can be flattened and passed through a deep neural network consisting of fully connected layers. The output of the deep neural neural network is a vector of pose parameter $\bm{p} \in \mathbb{R}^{72}$. The pose parameter can be used for a deformable human model for demonstrating a specific pose of the human, i.e., a standing pose with raised hands. The deformable human model is a parametric model, i.e., SMPL \cite{loper2023smpl}, that takes the pose and shape parameters as inputs and produces the outputs as high dimensional triangle meshes in 3D space, i.e., $\mathbf{M} \in \mathbb{R}^{6890 \times 3}$. In our paper, we only consider the pose parameter $\bm{p}$ as the input of the SMPL model and use a fixed shape parameter. 

In the online signal detection and inference phase, i.e., Fig~\ref{fig:imu-receiver}(b), raw IMU signals are quantized, modulated, and transmitted over an OFDM channel. The received OFDM symbols are decoded to binary digits by the pre-trained neural receiver. After that, a de-quantization step is performed to reconstruct $\bm{\hat{x}}$. The pre-trained IMU receiver then predicts the pose parameters $\bm{\hat{p}}$, which are used for real-time human pose reconstruction. Unlike the offline phase, no backpropagation occurs in this stage, allowing efficient and low-latency inference. Note that the main difference between our proposed IMU receiver and other methods in the literature, e.g., \cite{von2017sparse, huang2018deep} and \cite{hieu2024reconstructing}, is that the IMU receiver works under practical settings with distorted IMU signals due to transmission over the OFDM channel.

To train the IMU receiver, we use the DIP-IMU dataset that consists of a large IMU reading from classes of various human activities \cite{huang2018deep}. The deep neural network, i.e., Multi-Layer Perceptron (MLP), has one input layer of 204 units, two hidden layers each having 500 units, and the last output layer has 72 units. 
The IMU receiver is trained in a supervised manner using paired input-output samples \(\{\bm{x}^{(i)}, \bm{p}^{(i)}\}_{i=1}^N\), where \(\bm{x}^{(i)} \in \mathbb{R}^{204}\) denotes the flattened IMU sensor measurements, and \(\bm{p}^{(i)} \in \mathbb{R}^{72}\) represents the ground truth SMPL pose parameters. The training objective is to minimize the MSE between the predicted pose parameters \(\hat{\bm{p}}^{(i)}\) and the ground truth \(\bm{p}^{(i)}\). The loss function is defined as:  
\begin{equation}
\mathcal{L}_{\text{MSE}} = \frac{1}{N} \sum_{i=1}^{N} \left\| \boldsymbol{p}^{(i)} - \hat{\boldsymbol{p}}^{(i)} \right\|_2^2,
\label{eq:loss-mse}
\end{equation}
where \(\hat{\boldsymbol{p}}^{(i)} = f_{\bm{\Phi}}(\boldsymbol{x}^{(i)})\) is the output of the MLP parameterized by weights \(\bm{\Phi}\). The Adam optimizer is employed to update \(\bm{\theta}\) by backpropagating gradients of \(\mathcal{L}_{\text{MSE}}\) with a learning rate of \(0.001\), trained over \(50\) epochs. This ensures the model learns to map noisy or quantized IMU signals to accurate 3D poses, even under wireless channel distortions. Detailed parameters for the IMU receiver are shown in Table \ref{tab:simulation-params}.

Overall, the proposed framework integrates quantized IMU transmission over a ray-traced OFDM channel, followed by neural signal decoding and pose reconstruction. By jointly optimizing channel estimation and demodulation, the neural receiver minimizes BER, while the IMU receiver maps decoded bits to 3D poses using a pre-trained MLP. The training process of the IMU receiver is described in Algorithm \ref{algo:imu-receiver} \footnote{
Some general-purpose notations for hyper-parameters of the Algorithms and \ref{algo:neural-receiver} and \ref{algo:imu-receiver} are used interchangeably to avoid notation's complexity. For the detailed values of these parameters, please refer to Table \ref{tab:simulation-params}.
}.
Algorithm \ref{algo:imu-receiver} outlines the training process for the IMU receiver, which maps quantized IMU signals to 3D human poses. 
The algorithm begins by initializing the IMU receiver model $f_{\bm{\Theta}}$ with random weights $\bm{\Theta}$. For each epoch, the training dataset $D_{\text{train}}$ is shuffled to ensure stochasticity, followed by batch-wise processing. In each iteration, a batch of IMU signals $\bm{x}_{\text{batch}}$ is fed into the model to predict pose parameters $\bm{\hat{p}}_{\text{batch}}$. The MSE loss between predicted and ground-truth poses is computed, and gradients are backpropagated to update the model weights using the Adam optimizer with learning rate $\eta$. This iterative process continues for $N_{\text{epoch}}$ epochs, yielding a trained IMU receiver, which is capable of reconstructing 3D human poses from distorted de-quantized IMU signals.

\section{Simulation Results}
\label{sec:simulation}
\subsection{Channel Dataset Generation}

\begin{table*}[htbp]
\centering
\caption{Simulation Parameter Summary}
\label{tab:simulation-params}
\begin{tabular}{lll}
\toprule
\textbf{Category} & \textbf{Parameter} & \textbf{Value/Configuration} \\
\midrule
\multirow{8}{*}{OFDM System} 
& Carrier Frequency & 3.5 GHz \\
& Number of Subcarriers & 128 \\
& Subcarrier Spacing & 30 kHz \\
& Number of OFDM Symbols & 14 \\
& Guard Carriers & 5 (left), 6 (right) \\
& Pilot Configuration & Kronecker pattern 2P and 1P configurations (see Fig.~\ref{fig:pilot-config}) \\
& Modulation Scheme & 4-QAM  \\
& MIMO Configuration & 16 RX antennas (base station), 1 TX antenna (user) \\
\midrule
\multirow{14}{*}{Neural Receiver} 
& Architecture & Residual Convolutional Neural Network (ResNet) \\
& Input & Real/imaginary parts of received OFDM resource grid + noise variance \\
& Input Shape & \texttt{[100, 16, 14, 128]} \cite{honkala2021deeprx} \cite{aoudia2021end} \\
& & (Batch size \(\times\) RX antennas \(\times\) OFDM symbols \(\times\) Subcarriers) \\
& Residual Blocks & 4 blocks, each with: \\
& & - 2 Conv2D layers (128 filters, 3\(\times\)3 kernel, ReLU activation) \\
& & - Layer normalization over time, frequency, and channel dimensions \\
& & - Skip connection \\
& Output Conv & Conv2D (2 filters, 3\(\times\)3 kernel, linear activation) \\
& Output & Log-Likelihood Ratios (LLRs) for decoded bits \cite{aoudia2021end} \\
& Loss Function & Binary Cross-Entropy \\
& Optimizer & Adam \\
& Learning Rate & 0.001 \\
& Training Epochs & 100,000 \\
\midrule
\multirow{7}{*}{IMU Receiver} 
& Architecture & Multi-Layer Perceptron (MLP) \\
& Input & Flattened IMU data (204 features) \\
& Hidden Layers & 3 layers: 512 \(\rightarrow\) 256 \(\rightarrow\) 128 units (ReLU activation) \\
& Output Layer & 72 units (linear activation, SMPL pose parameters) \cite{huang2018deep} \cite{hieu2024reconstructing} \\
& Loss Function & Mean Squared Error (MSE) \\
& Optimizer & Adam \\
& Learning Rate & 0.001 \\
& Training Epochs & 50 \\
\midrule
\multirow{2}{*}{Quantization} 
& IMU Data Precision (Original) & 32-bit floating point \\
& Quantization Levels (\(q\)) & \(q = 4, 5, 6, 7, 8, 9, 10\) bits \\
\midrule
\multirow{8}{*}{Channel Dataset} 
& \(E_b/N_0\) Range & \(-5.0\ \text{dB}\) to \(16.0\ \text{dB}\) \cite{aoudia2021end} \\
& Number of CIRs & 11,796 \\
& Number of 3D Maps & 2 (Munich and Detoile) \\
& Ray Tracing Depth	 & 5 (maximum reflections/refractions) \\
& Ray Tracing Paths & 75 paths (max) \\
& Path Gain Range & -130 dB (min) to 0 dB (max) \\
& Transmitter Velocity & [13.6–18.8,13.6–18.8,0.0] m/s (randomized x/y motion) \cite{aoudia2021end} \\
& Doppler Sampling Frequency & 30 kHz \\
\midrule
\multirow{5}{*}{IMU Dataset} 
& IMU Sensors per User & 17 Xsens sensors \cite{xsens2023} \cite{huang2018deep}\\
& Training Dataset's Size & 20,076 \\
& Test Dataset's Size & 6,990 \\
& Sampling Rate & 60 Hz (60 IMU frames per second) \\
& Human Poses & Upper body, lower body, locomotion, freestyle, and interaction \cite{huang2018deep}  \\
\bottomrule
\end{tabular}
\end{table*}

The generation of the OFDM channel dataset is crucial for training the proposed neural receiver. As discussed in Section \ref{subsec:rt-introduction}, we employ a site-specific ray tracing approach to model the wireless propagation environment, capturing realistic multipath effects, spatial characteristics, and Doppler shifts. This ensures that the neural receiver is trained on data that accurately reflects real-world wireless conditions rather than relying on conventional statistical fading models.
To construct the dataset, we utilize two representative 3D urban environments, namely \textit{Munich} and \textit{Etoile}, using Sionna's ray tracing toolkit \cite{sionna}. These environments contain detailed building layouts and material properties that influence signal propagation. The receiver (base station) is deployed at a fixed position with a carrier frequency of {3.5}{GHz}, while the transmitter (user) is placed at various positions sampled from a coverage map within a radius of \SI{10}{m} to \SI{400}{m} around the base station. The user moves with a velocity sampled uniformly between \SI{13.6}{m/s} and \SI{18.8}{m/s} to simulate realistic user mobility in urban environments \cite{aoudia2021end}.

Each channel realization is derived from a ray tracing simulation that determines the number of propagation paths, their corresponding complex gains, and delays. The CIR extracted from the ray tracing model consists of up to \( 75 \) propagation paths, each with a path gain \( a_i \) and delay \( \tau_i \). These are then transformed into the OFDM frequency response using the system parameters defined in Table I\ref{tab:simulation-params}. Specifically, for an OFDM system with \( N = 128 \) subcarriers and subcarrier spacing \( \delta f = \SI{30}{kHz} \), the channel frequency response at the \( n \)-th subcarrier is:

\begin{equation}
    H[n] = \sum_{i=1}^{L} a_i e^{-j 2\pi (f_c + n\delta f) \tau_i}, \quad n \in \{-64, \dots, 63\}.
\end{equation}

This transformation allows for direct integration of the ray tracing channel model into the neural receiver's training dataset.
To account for the effects of receiver mobility, Doppler shifts are introduced using a Doppler sampling frequency of \SI{30}{kHz}. The frequency shift experienced by each multipath component depends on the user's velocity and the AoA, ensuring an accurate representation of time-varying fading effects. 

The final dataset consists of \( 11,796 \) CIRs, which will be further split into a training set and a test set with a ratio of 80:20, respectively. The dataset is stored as structured arrays to enable efficient batch processing during neural receiver training. By leveraging ray tracing-based modeling, our approach allows the neural receiver to learn robust channel estimation and decoding strategies under diverse site-specific conditions, bridging the gap between theoretical OFDM models and real-world wireless environments. Detailed parameter settings for the channel dataset can be found in Table \ref{tab:simulation-params}.

\subsection{Bit Error Rate Performance}
\begin{figure}
\includegraphics[width=0.9\linewidth]{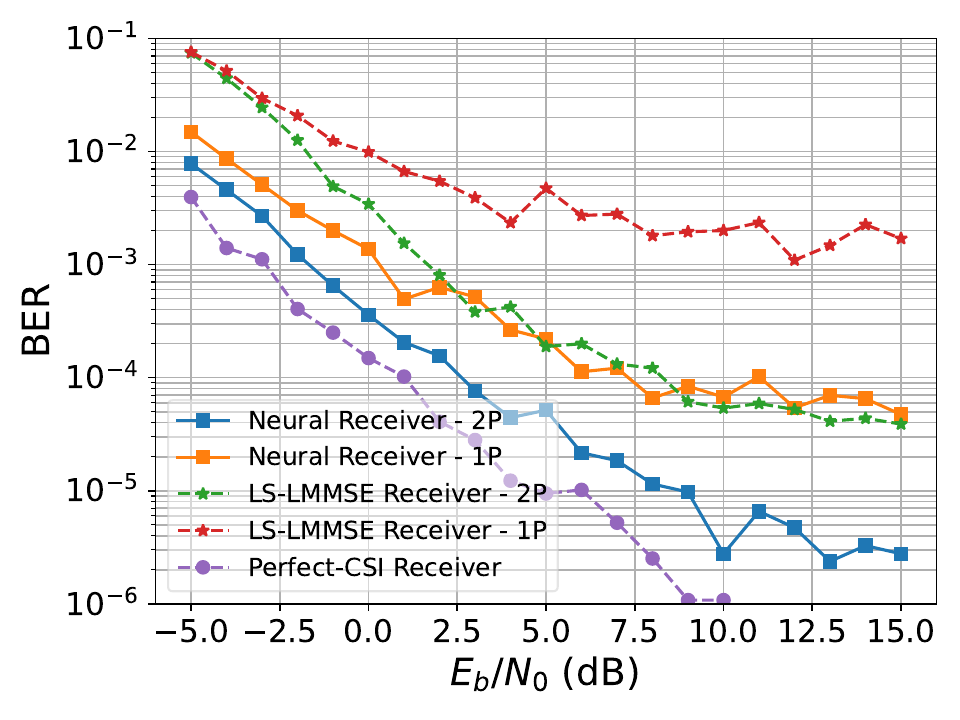}
\caption{Bit error rate values under various $E_b/N_0$ conditions.}
\label{fig:ber}
\end{figure}

We first evaluate the performance of the neural receiver on the conventional BER performance. The BER performance of the neural receiver will directly impact the performance of the IMU receiver in the later stage. 
The BER performance of the proposed neural receiver scheme is illustrated in Fig.~\ref{fig:ber}. Note that the results are obtained for two scenarios 2P and 1P, reflecting different configurations of pilot symbols in the OFDM resource grid as illustrated in Fig.~\ref{fig:pilot-config}. We only report one result for the perfect channel state information (perfect CSI) scenario as the results for 2P and 1P configurations are identical in this scenario. As observed in Fig.~\ref{fig:ber}, the perfect CSI scenario achieves the lowest BER values across the range of $E_b/N_0$ (i.e., signal power to noise power ratio). This is because, in the perfect CSI scenario, the receiver is assumed to know the channel matrix $\mathbf{H}_{ij}$ in advance, yielding error-free channel estimation. 

In the 2P scenario, we observe the performance gap between the proposed neural receiver scheme (blue line) and the LS-LMMSE scheme (dashed green line). For example, the gap between the neural receiver and the LS-LMMSE receiver is 5 dB at $10^{-4}$ BER. This is because the end-to-end learning scheme used for the neural receiver benefits from the supervised training, thus obtaining lower estimation errors propagating through the signal processing stages compared to the conventional receiver with LS channel estimation and LMMSE equalization. A similar performance gap can be observed in the 1P scenario. However, in the 1P scenario, the neural receiver and LS-LMMSE receiver achieve higher BER values, compared to their 2P counterparts. The reason for the higher BER values in the 1P scenario relies on the lower number of pilot symbols in the OFDM resource grid. As observed in Fig.~\ref{fig:pilot-config}, the 1P scenario only contains one column of pilot symbols at the third OFDM symbol, making the signal interpolation using LMMSE equalization less accurate. In contrast, the 2P scenario has two OFDM symbol slots for the pilot symbols, making the interpolation of the estimated symbols from pilot symbol positions more accurate, regardless of the effects of the Doppler shift. 

\subsection{Impacts of Quantization}
\begin{figure}
\includegraphics[width=0.9\linewidth]{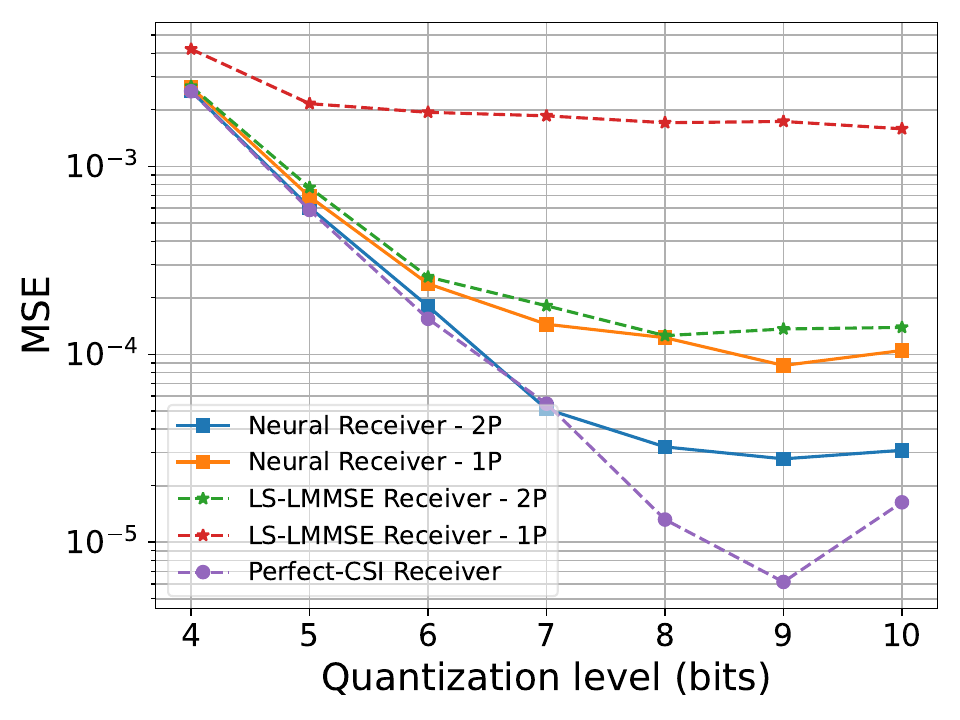}
\caption{Mean squared error under different quantization levels at $E_b/N_0 = 5.0$ dB.}
\label{fig:mse}
\end{figure}

\begin{figure*}[t]
\centering
\includegraphics[width=0.8\linewidth]{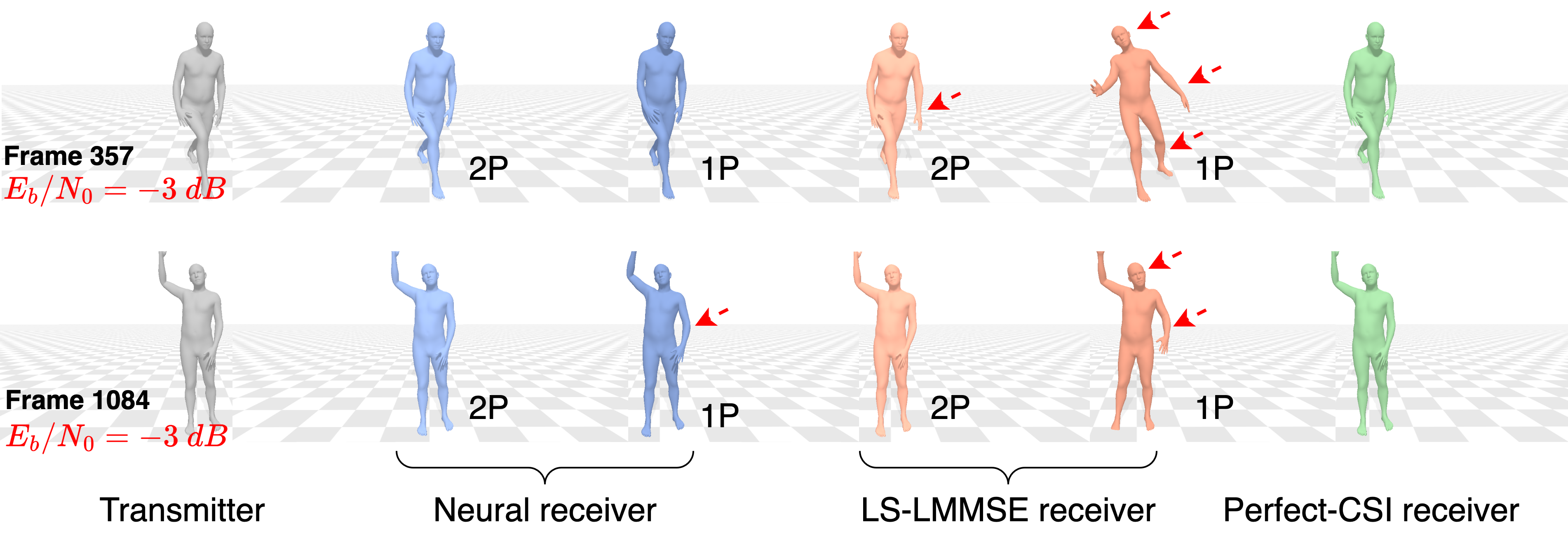}
\caption{Human body movement reconstructed at the receiver (base station).}
\label{fig:pose-2p1p}
\end{figure*}

\begin{figure*}[t]
\centering
\includegraphics[width=0.7\linewidth]{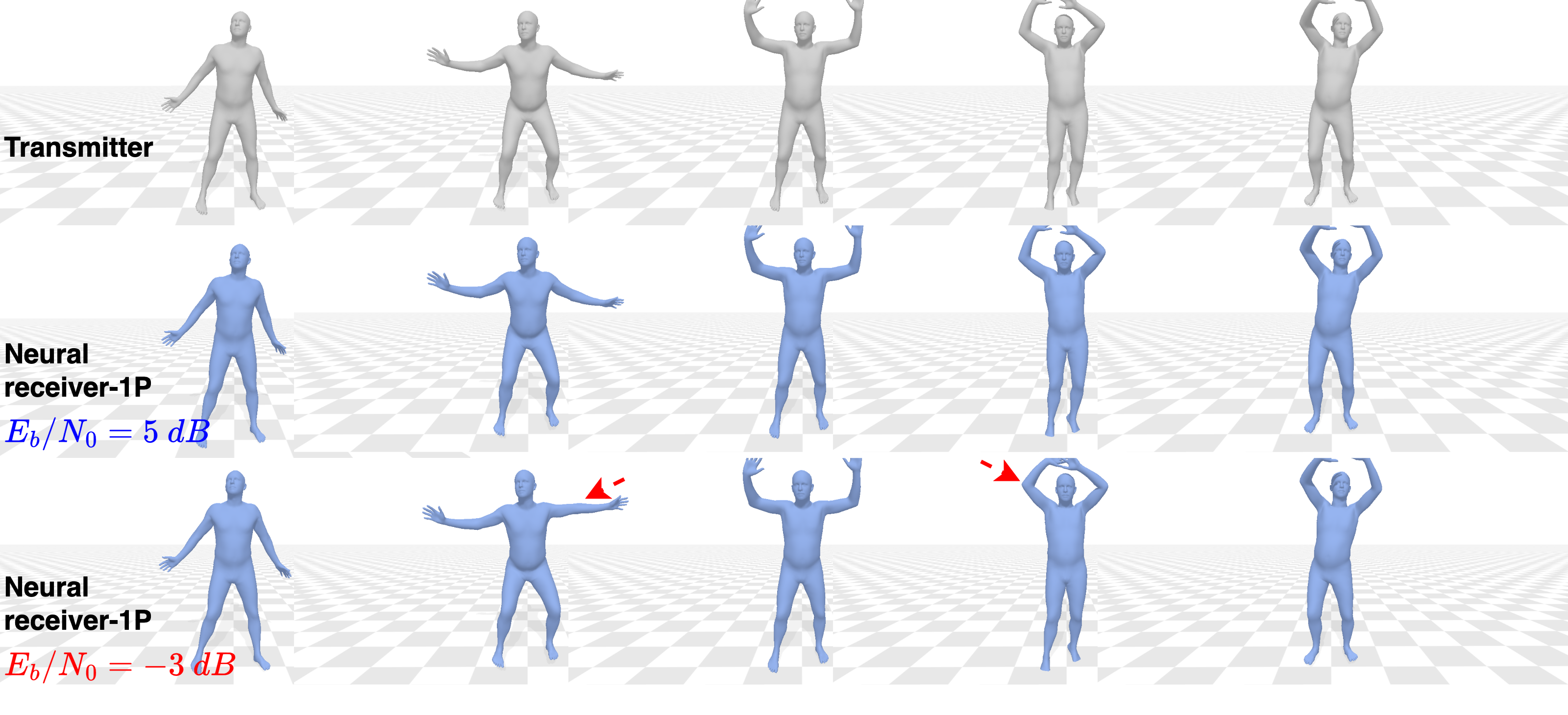}
\caption{Animation of body movement reconstructed at the receiver with $E_b/N_0 = 5$ dB and $E_b/N_0 = -3$ dB.}
\label{fig:pose-ebno}
\end{figure*}

\begin{figure}[t]
\centering
\includegraphics[width=0.9\linewidth]{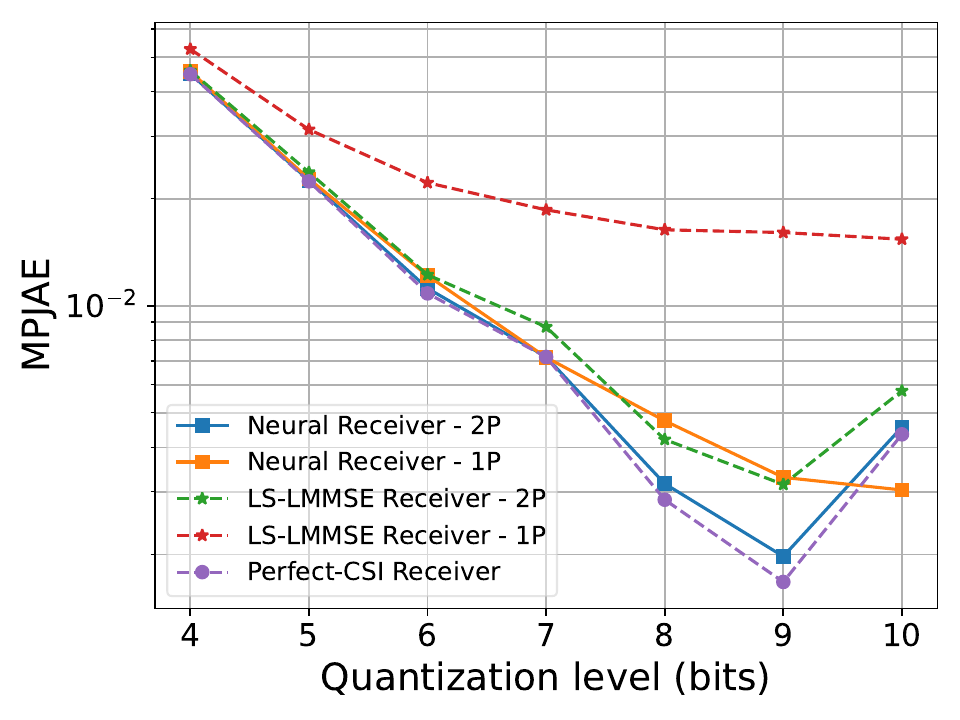}
\caption{Mean per joint angular error ($^{\circ}$) of the reconstructed poses compared to the ground truth values. The results are obtained at $E_b/N_0 = 5.0$ dB.}
\label{fig:mpjae}
\end{figure}

Next, we evaluate the impacts of quantization on the performance of the reconstructed human pose in Fig.~\ref{fig:mse}. In Fig.~\ref{fig:mse}, we evaluate the MSE value between the original IMU signals and the received IMU signals with the presence of quantization.
We observe that the increase of the quantization level, i.e., $q$ in equation (\ref{eq:quantization}), results in the decrease of the MSE values for all the receivers. Notable, the decreasing MSE values saturate from the quantization level $q=8$ for most of the schemes. This suggests that the use of full 32 floating bits to transmit the IMU signals is redundant as the higher quantization levels, i.e., $q \geq 8$, do not result in lower MSE values. In Fig.~\ref{fig:mse}, we can observe the performance gap between the neural receiver and LS-LMMSE receiver in scenarios 2P and 1P, respectively. For example, the neural receiver (2P) achieves $3 \times 10^{-5}$ MSE value while the LS-LMMSE receiver (2P) only obtains $1.3 \times 10^{-4}$ MSE value at quantization levels from 8 to 10. The larger performance gap between the neural receiver and the LS-LMMSE receiver in the 1P scenario also suggests that using more pilot symbols is more reliable for transmitting our IMU signals.

\subsection{Pose Reconstruction Evaluation}

Next, we use the IMU receiver to map the received IMU signals to the 3D human poses and illustrate the results in Fig.~\ref{fig:pose-2p1p} and Fig.~\ref{fig:pose-ebno}. In Fig.~\ref{fig:pose-2p1p}, we illustrate the performance gaps between the MSE values of the proposed neural receiver to the other baselines. The two rows of Fig.~\ref{fig:pose-2p1p} illustrate the ground truth human pose (denoted as ``Transmitter") and the human poses reconstructed by the receivers in two different IMU frames. The evaluated $E_b/N_0$ is $-3$ dB. We observe that the human pose reconstructed at the neural receiver in the 1P scenario experiences a slight error in left arm movement, marked with the red arrow. The results obtained by the LS-LMMSE receiver are worse than those of the neural receiver as there are more error movements in the human poses. The perfect CSI scenario achieves the highest human pose accuracy as the receiver in this scenario does not experience channel estimation errors. 

Next, we evaluate the impacts of $E_b/N_0$ values on the reconstructed human poses at the receiver in Fig.~\ref{fig:pose-ebno}. In Fig.~\ref{fig:pose-ebno}, the first row is the ground truth poses, and the second and third rows are the human poses reconstructed at different $E_b/N_0$ values. The first row (denoted as ``Transmitter") illustrates 6 frames of the user movement in a jumping jack data sequence. We select the jumping jack movement of the user from the test set because this is a challenging task with high-frequency movements of arms and legs. Notably, the results of the proposed neural receiver with 1P configuration and $E_b/N_0 = 5$ dB are identical to the ground truth. This suggests that the neural receiver can reconstruct high-frequency movements of the user from IMU signals with high accuracy. In the $E_b/N_0 = -3$ dB scenario, we observe some slight errors in hand movements, marked with red arrows. More visualization results can be found on our project page in \url{https://github.com/TheOpenSI/imu2pose-sionna}.   

Finally, to further analyze the robustness of the reconstructed human poses, Fig.~\ref{fig:mpjae} presents the Mean Per Joint Angular Error (MPJAE), which quantifies the average angular difference (in degrees) between the predicted and ground-truth joint rotations. Given the SMPL pose parameters, where each joint rotation is represented as an axis-angle vector, MPJAE is defined as:

\begin{equation}
    \text{MPJAE} = \frac{1}{K} \sum_{i=1}^{K} \left\| \hat{\omega}_i - \omega_i \right\|_1,
\end{equation}
where $K=24$ is the total number of SMPL joints, $\omega_i \in \mathbb{R}^{1 \times 3}$ and $\hat{\omega}_i \in \mathbb{R}^{1 \times 3}$ are the ground truth and reconstructed joint rotation vectors (3D axis-angle representations), which can be directly obtained by extracting from the SMPL pose parameters $\bm{p} \in \mathbb{R}^{72}$ and $\hat{\bm{p}} \in \mathbb{R}^{72}$, respectively. 

Note that lower MPJAE values indicate higher pose reconstruction accuracy. The results show that increasing the quantization level consistently reduces MPJAE, highlighting that finer quantization improves the precision of the reconstructed poses. However, beyond $q = 8$ bits, the accuracy gain plateaus, confirming that additional precision offers diminishing returns, as also observed in Fig.~\ref{fig:mse}. We observe the slight fluctuation of the MPJAE values at quantization levels 9 and 10 because of the inconsistency in user movements. For example, high-frequency movements such as the jumping jack in Fig.~\ref{fig:pose-ebno} can result in changing MSE values, yielding fluctuated MPJAE values as observed in Fig.~\ref{fig:mpjae}. Nevertheless, with finer quantization levels $q \geq 7$, we observe that the MPJAE values of the proposed approaches are less than $0.01^{\circ}$, indicating robust reconstructed poses \cite{von2017sparse, huang2018deep}. The results show a reduction of $37\%$ in MPJAE value by using the neural receiver (2P) compared to the baseline LS-LMMSE receiver (2P) at 9-bit quantization. 
In addition, the neural receiver outperforms the LS-LMMSE receiver across all quantization levels, with a more significant advantage in the 1P configuration, demonstrating its robustness under reduced pilot density. The Perfect-CSI receiver baseline achieves the lowest MPJAE, reaffirming that eliminating channel estimation errors leads to near-ideal pose recovery. These results further validate the effectiveness of the proposed neural receiver in achieving high-fidelity human pose reconstruction under practical wireless conditions.

\section{Conclusion}
\label{sec:conclusion}
In this work, we have developed a novel deep learning-based framework for 3D human pose reconstruction from IMU signals over an OFDM system. Unlike traditional approaches that assume ideal transmission conditions, our framework explicitly accounts for realistic wireless impairments, including bit errors and quantization effects. By integrating a neural receiver at the base station, we jointly optimize OFDM channel estimation and signal decoding, significantly improving the BER performance over conventional LS-LMMSE receivers.
A key contribution of this work is the demonstration that low-bit quantization (e.g., 8-bit precision) is sufficient for accurate human pose reconstruction, eliminating the need for high-precision (32-bit) raw IMU transmissions. Our experimental results show that, at $E_b/N_0 = 5$ dB, the proposed neural receiver effectively mitigates channel-induced errors, achieving smooth and realistic human pose animations even under challenging conditions such as high-motion activities.
Through extensive simulations using a site-specific OFDM channel model generated via ray tracing, we validate the practical feasibility of our approach in future 6G extended reality (XR) applications. Our findings highlight the critical interplay between wireless channel reliability, signal quantization, and human pose fidelity, paving the way for seamless XR experiences over next-generation wireless networks.
Future research directions include extending our framework to multi-user scenarios and incorporating adaptive modulation and coding schemes. Additionally, investigating the trade-off between latency and pose reconstruction accuracy will be crucial for enabling real-time XR applications.

\bibliographystyle{IEEEtran}
\bibliography{refs}
\end{document}